\documentclass[runningheads,a4paper]{llncs}

\usepackage{amssymb}
\usepackage{graphicx}

\usepackage{url}
\urldef{\mailsa}\path|{souryade, yinansha, chugg, pabeerel}@usc.edu|    
\newcommand{\keywords}[1]{\par\addvspace\baselineskip
\noindent\keywordname\enspace\ignorespaces#1}

\makeatletter
\def\blfootnote{\xdef\@thefnmark{}\@footnotetext}
\makeatother

\begin{document}

\mainmatter

\blfootnote{The final publication is available at Springer via\\ \url{http://dx.doi.org/10.1007/978-3-319-68600-4_32}}

\title{Accelerating Training of Deep Neural\\Networks via Sparse Edge Processing}

\titlerunning{Accelerating Training of DNNs via Sparse Edge Processing}

\author{Sourya Dey\and Yinan Shao\and Keith M.~Chugg\and Peter A. Beerel}

\authorrunning{Sourya Dey, Yinan Shao, Keith M.~Chugg, and Peter A. Beerel}

\institute{Ming Hsieh Department of Electrical Engineering,\\
University of Southern California.\\
Los Angeles, California 90089, USA\\
\mailsa}

\toctitle{Accelerating Training of DNNs via Sparse Edge Processing}
\tocauthor{Sourya Dey}
\maketitle

\begin{abstract}
We propose a reconfigurable hardware architecture for deep neural networks (DNNs) capable of online training and inference, which uses algorithmically pre-determined, structured sparsity to significantly lower memory and computational requirements. This novel architecture introduces the notion of edge-processing to provide flexibility and combines junction pipelining and operational parallelization to speed up training. The overall effect is to reduce network complexity by factors up to 30x and training time by up to 35x relative to GPUs, while maintaining high fidelity of inference results. This has the potential to enable extensive parameter searches and development of the largely unexplored theoretical foundation of DNNs. The architecture automatically adapts itself to different network sizes given available hardware resources. As proof of concept, we show results obtained for different bit widths.
\end{abstract}

\section{Introduction}\label{intro}

DNNs in machine learning systems are critical drivers of new technologies such as natural language processing, autonomous vehicles, and speech recognition.
Modern DNNs and the corresponding training datasets are  gigantic 
with millions of parameters \cite{Krizhevsky2012}, which makes training a painfully slow and memory-consuming experimental process. For example, one of the winning entries in the ImageNet Challenge 2014 takes 2-3 weeks to train on 4 GPUs \cite{Simonyan2014}. As a result, despite using costly cloud computation resources, training is often forced to exclude large scale optimizations over model structure and hyperparameters.
This scenario severely hampers the advancement of research into the limited theoretical understanding of DNNs and, unfortunately, empirical optimizations remain as the only option.

Recent research into 
hardware architectures for DNNs has primarily focused on inference only, while performing training \emph{offline} \cite{Chen2014DN,Chen2014DDN,Han2016EIE,Himavathi2007,Sanni2015,Zhang2016} .
Unfortunately, this precludes reconfigurability and results in a network incapable of dynamically adapting itself to new patterns in data, which severely limits its usability for pertinent real-world applications such as stock price prediction and spam filtering. Moreover, offline-only learning exacerbates the problem of slow DNN research and ultimately leads to lack of transparency at a time when precious little is understood about the working of DNNs.

There has been limited research into hardware architectures to support \emph{online} training, such as \cite{Ahn2014,Eldredge1994,Gadea2000}.  
However, due to the space-hungry nature of DNNs, these works have only managed to fit small networks on their prototypes. While other works \cite{Chen2015,Han2016EIE,Han2016DC,Han2015,Zhou2016} have proposed 
memory-efficient solutions for inference, 
none of them have addressed the cumbersome problem of online training. Therefore, a hardware architecture supporting online training and reconfiguration of large networks would be of great value for exploring a larger set of models for both empirical optimizations and enhanced scientific understanding of DNNs.

In this work, we propose a novel hardware architecture for accelerating training and inference of DNNs on 
FPGAs. Our key contributions are:
\begin{enumerate}
    \item An architecture designed for FPGA implementation that can perform online training of large-scale DNNs.
    \item A pre-defined, structured form of sparsity that starts off with an algorithmically deterministic sparse network from the very outset.
    \item Edge-based processing -- a technique that decouples the available hardware resources from the size and complexity of the network, thereby leading to tremendous flexibility and network reconfigurability.
    \item Hardware-based optimizations such as operational parallelization and junction pipelining, which lead to large speedups in training.
\end{enumerate}

The paper is organized as follows. Section \ref{sparsity} analyzes our proposed form of sparsity. Section \ref{edges} discusses our proposed technique of edge-based processing and interleaving, along with hardware optimizations. Then section \ref{results} presents hardware results and section \ref{future} concludes the paper.

\section{Sparsity}\label{sparsity}

The need for sparsity, or reducing the number of parameters in a network, stems from the fact that both the memory footprint and computational complexity of modern DNNs is enormous. For example, the well-studied DNN AlexNet \cite{Krizhevsky2012} has a weight size of 234 MB and requires 635 million arithmetic operations only for feedforward processing \cite{Zhang2016}. Convolutional layers 
are sparse, but \emph{locally connected}, i.e. the spatial span of neurons in a layer connecting to a neuron in the next layer is small. As a result, such layers alone are not suitable for performing inference and therefore need to be followed by fully-connected (FC) layers \cite{Krizhevsky2012,Simonyan2014,Szegedy2014}, which account for 95\% of the connections in the network \cite{Zhang2016}. However, FC layers are typically over-parameterized \cite{Cun1990,Denil2013} and tend to \emph{overfit} to the training data, which results in inferior performance on test data. Dropout (deletion) of random neurons was proposed by \cite{Srivastava2014}, but incurs the disadvantage of having to train multiple differently configured networks, which are finally combined to regain the original full size network. Hashnet \cite{Chen2015} randomly forced the same value on collections of weights, but acknowledged that ``a significant number of nodes [get] disconnected from neighboring layers.'' Other sparsifying techniques such as pruning and quantization \cite{Han2016DC,Han2015,Zhou2016} first train the complete network, and then perform further computations to delete parameters, which increase the training time. In general, all of these architectures deal with the complete non-sparsified FC layers at some point of time during their usage cycle and therefore, fail to permanently solve the memory and complexity bottlenecks of DNNs. 

Contrary to existing works, we propose a class of DNNs with \emph{pre-specified sparsity}, implying that from the very beginning, neurons in a layer connect to only a subset of the neurons in the next layer. This means that the original network has a lower memory and computational complexity to begin with, and there are no 
additional computations to change the network structure.
The degrees of \emph{fan-out} and \emph{fan-in} (number of connections to the next layer and from the previous layer, respectively) of each neuron are user-specified, and then the connections are algorithmically assigned. This ensures that no particular neuron gets disconnected, while the algorithm provides good \emph{spatial spread} ensuring that activations from early layers can impact the output of the last layer.

\begin{figure}
\centering
\includegraphics[width = 0.58\linewidth]{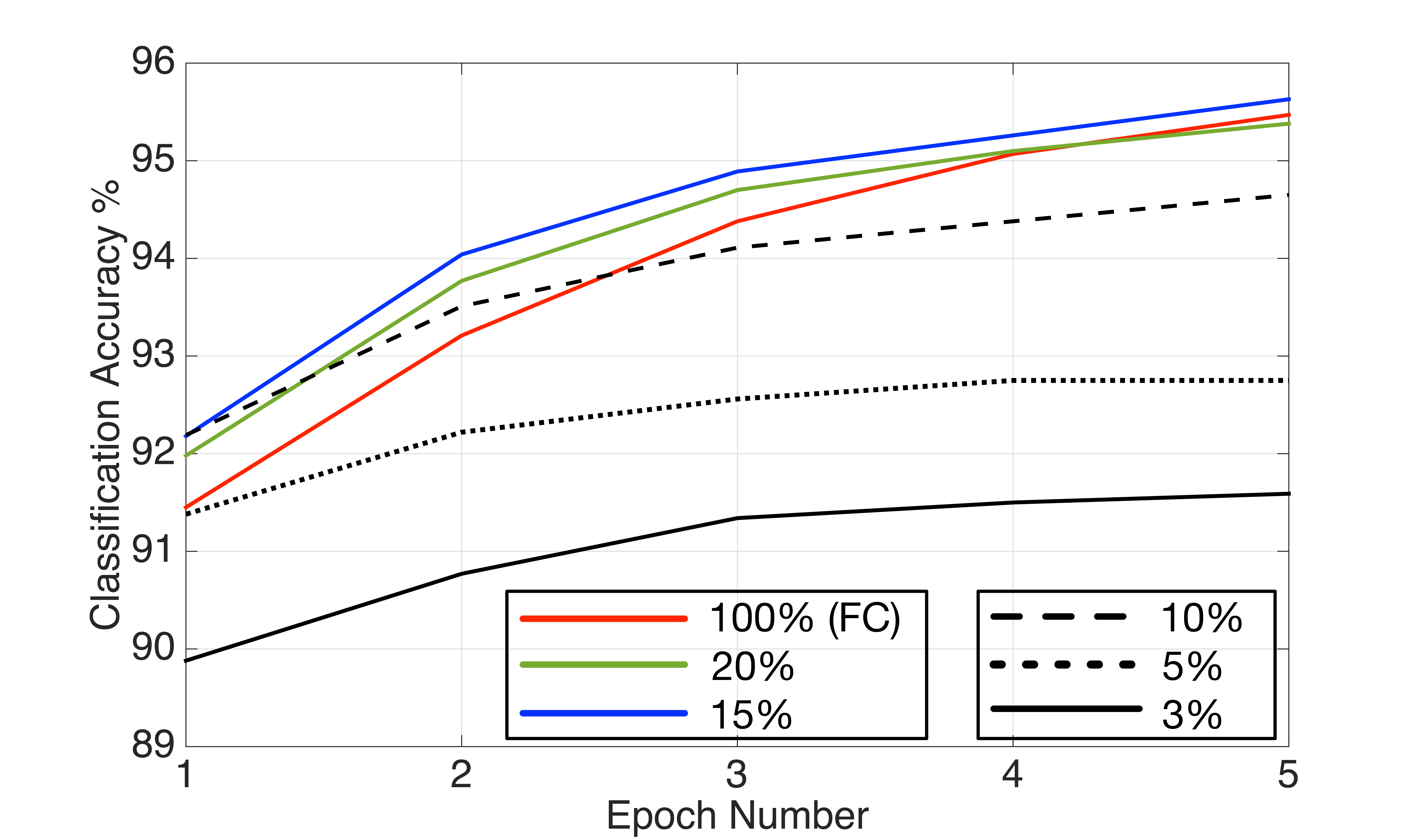}
\caption{Classification performance of a (784,112,10) network with varying connectivity percentage, trained for 5 epochs on the MNIST dataset. 
}
\label{fig-sparsity}
\end{figure}

As an example, consider MNIST digit classification over 5 epochs of training using a (784,112,10) network, i.e. there are 784 input, 112 hidden and 10 output neurons. If it is FC, the total number of weights is
88,928 (which is already less than other works such as \cite{Ciresan2010}). Now suppose we preset the fan-out of the input and hidden neurons to 17 and 5, respectively. This 
leads to 13,888 total weights, implying that the overall network has 15\% connectivity, 
or 85\% sparsity. Figure \ref{fig-sparsity} compares the performance of sparse networks, keeping all hyperparameters the same except for adjusting the learning rate to be inversely proportional to connectivity, which compensates for parameter reduction. 
Notice that 15\% connectivity gives better performance than the original FC network. Moreover, 3\% connectivity gives $>91\%$ accuracy in 5 epochs, which is within 4\% of the FC case. This leads us to believe that the memory and processor requirements of FC layers in  DNNs can be reduced by over 30x with minimal impact on performance.

\section{Edge Processing and Interleaving}\label{edges}

A DNN is made up of layers of interconnected neurons and the \emph{junctions} between adjacent layers contain connections or \emph{edges}, each having an associated \emph{weight} value. The 3 major operations in a network are: a) feedforward (FF), which primarily computes dot products between weights and the previous layer's activation values; b) backpropagation (BP), which computes dot products between weights and the next layer's delta values and then multiplies them with derivatives of the previous layer's activations; and c) update (UP), which multiplies delta values from the next layer with activation values from the previous layer to compute updates to the weights. Notice that the edges  
feature in all 3 operations, and this is where the motivation for
our approach stems from.

We propose a DNN architecture which is processed from the point of view of its edges (i.e., weights), instead of its neurons. Every junction has a \emph{degree of parallelism (DoP)}, denoted as $z$, which is the \emph{number of edges processed in parallel}. All the weights in each junction are stored in a memory bank consisting of $z$ memories. All the activation, activation derivative and delta values of each layer are also stored in separate memory banks of $z$ memories each. The edges coming into a junction from its preceding layer are \emph{interleaved}, or permuted, before getting connected to its succeeding layer.
The interleaver algorithm is deterministic and reconfigurable. It serves to ensure good spatial spread and prevent regularity, thereby achieving a pseudo-random connection pattern. For example, if 4 edges come out of the first input neuron of the (784,112,10) network, they might connect to the 9th, 67th, 84th and 110th neuron in the hidden layer.

\begin{figure}
\centering
\includegraphics[width = 1.0\linewidth]{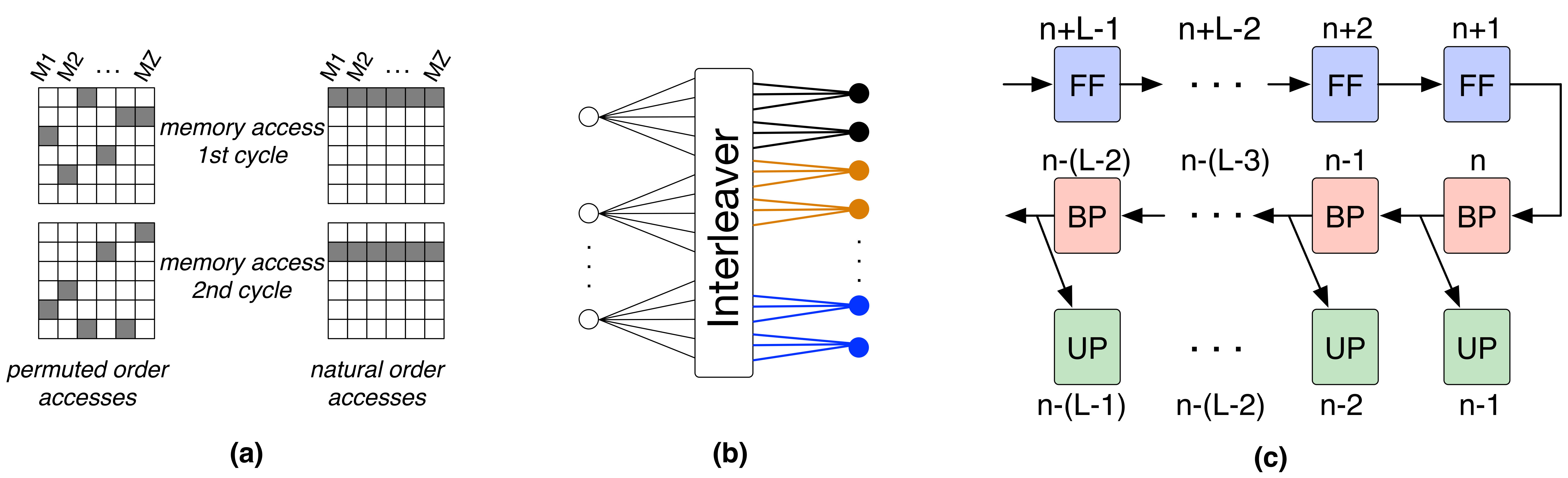}
\caption{\textbf{(a)}: \emph{Natural order} and \emph{permuted order} access of memory banks. \textbf{(b)}: Reading $z=6$ weights (corresponding to 2 succeeding layer neurons) in each cycle. \textbf{(c)}: Junction pipelining and operational parallelization in the whole network. }
\label{fig-DoP}
\end{figure}

Figure \ref{fig-DoP}a depicts a memory bank as a checkerboard, where each column is a memory. A single \emph{cycle} of processing (say the $n$th) comprises accessing the $n$th cell in each of the $z$ weight memories. This implies reading all $z$ values from the same row (the $n$th), which we refer to as \emph{natural order} access. Reading a row implies accessing weights of edges connected to consecutive neurons in the succeeding layer, since that's how they are numbered. Figure \ref{fig-DoP}b gives an example where $z$ is 6 and fan-in is 3. The interleaver determines which neurons in the preceding layer are connected to those $z$ edges. For ideal spatial spread, these will be $z$ different neurons. The interleaver algorithm is also designed to be \emph{clash-free}, i.e. it ensures that the activation values of these $z$ preceding neurons are stored in $z$ different memories. Violating this condition leads to the same memory needing to be accessed more than once in the same cycle, i.e. a clash, which stalls processing. A consequence of clash-freedom and pseudo-random connection pattern is that the activation memories are accessed in \emph{permuted order}, as shown in Figure \ref{fig-DoP}a, where there is only 1 shaded cell in each column.

\begin{figure}
\centering
\includegraphics[width = 0.9\linewidth]{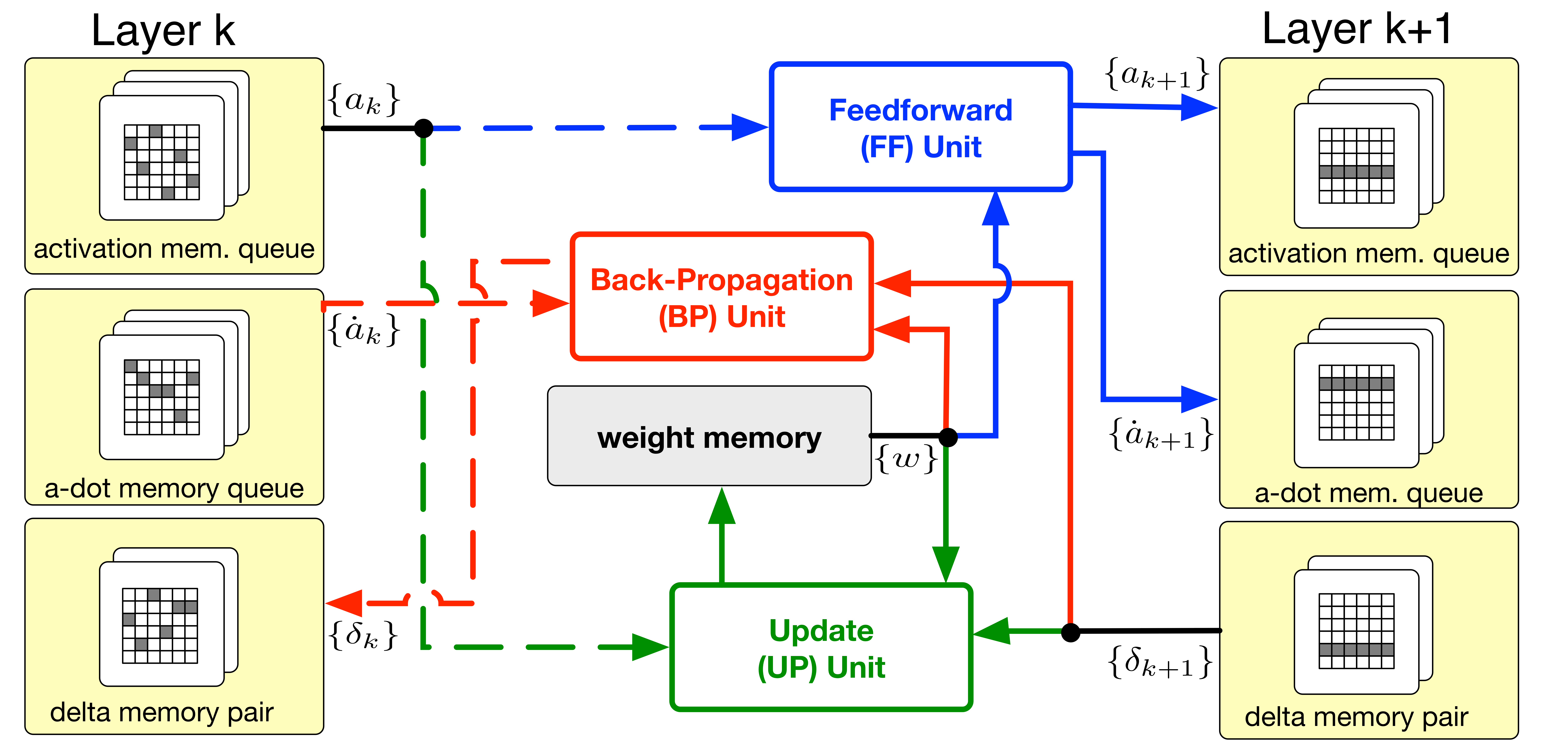}
\caption{Operational parallelization in a single junction between layers $k$ and $k+1$, showing natural and permuted order operations as solid and dashed lines, respectively.}
\label{fig-pnp}
\end{figure}

Noting the significant data reuse between FF, BP and UP, we used \emph{operational parallelization} to make all of them occur 
simultaneously. Since every operation in a junction uses data generated by an adjacent junction or layer, we designed a \emph{junction pipelining} architecture where all the junctions execute all 3 operations simultaneously on different inputs from the training set. This achieves a $3(L-1)$ times speedup, where $L$ is the total number of layers. The high level view is shown in Figure \ref{fig-DoP}c. As an example, consider the (784,112,10) network. When the second junction is doing FF on input $n+1$, it is also doing BP on the previous input $n$ which just finished FF, as well as updating (UP) its weights from the finished BP results of input $n-1$. Simultaneously, the first junction is doing FF on the latest input $n+2$, BP on input $n-1$, and UP using the BP results of input $n-2$. Figure \ref{fig-pnp} shows the 3 simultaneous operations in more detail inside a single junction. Notice that the memories associated with layer parameters are both read from and written into during the same cycle. Moreover, the activation and its derivative memories need to store the FF results of a particular input until it comes back to the same layer during BP. Hence these memories are organized in queues. 
While this increases overall storage space, the fraction is insignificant compared to the memory required for weights. 
This problem is alleviated by using only 1 weight memory bank per junction for all 3 processes. Moreover, only 2 rows of this bank need to be accessed at a time, 
which
makes efficient memory management techniques  possible.

A key contribution of our architecture is that $z$ can be set to any value depending on the area-speed tradeoff desired. $z$ can be made small to process a large network slowly using limited hardware. For powerful FPGAs, $z$ can be made large, which achieves tremendous increase in speed at the cost of a large number of multipliers. $z$ can also be individually adjusted for each junction so that the number of clock cycles to process each junction is the same, which ensures an always full pipeline and no stalls. Thus, the size and complexity of the network is decoupled from the hardware resources available. Moreover, low values of connectivity alleviate challenges with weight storage for very large DNNs. Our architecture can be reconfigured to varying levels of fan-out and structured sparsity, which is neither possible in other online learning architectures such as \cite{Ahn2014,Eldredge1994,Gadea2000}, nor in architectures using forms of unstructured sparsity that suffer from the overhead of lookups and cache misses \cite{Szegedy2014}. Thus, we achieve the ideal case of one-size-fits-all -- an architecture that can adapt to a large class of sparse DNNs.

As a concrete example of speedup, consider the network formed by the FC layers of AlexNet. This has a (1728,4096,4096,1000) neuron configuration and accounts for 6\% of the computational complexity \cite{Krizhevsky2012,Zhang2016}. Since the entire AlexNet takes 6 days to train on 2 GPUs for 90 epochs, we estimate that training only the FC network would take 0.36 days. The authors in \cite{Krizhevsky2012} acknowledge the over-parameterization problem, so we estimate from the data for Figure \ref{fig-sparsity} that the same FC network with only 6\% connectivity can be trained with minimal performance degradation. Using our architecture, modern Kintex Ultrascale FPGA boards will be able to support $z=256$. This results in 4096 cycles being needed to train a junction, which, at a reasonable clock frequency of 250 MHz, processes each image through this sparse FC network in 16 $\mu$s. Training the network for the complete 1.2 million images over 90 epochs is estimated to take half an hour, which is a speedup of 35x over a single GPU. 


\section{Results}\label{results}


\begin{figure}
\centering
\includegraphics[width = 0.9\linewidth]{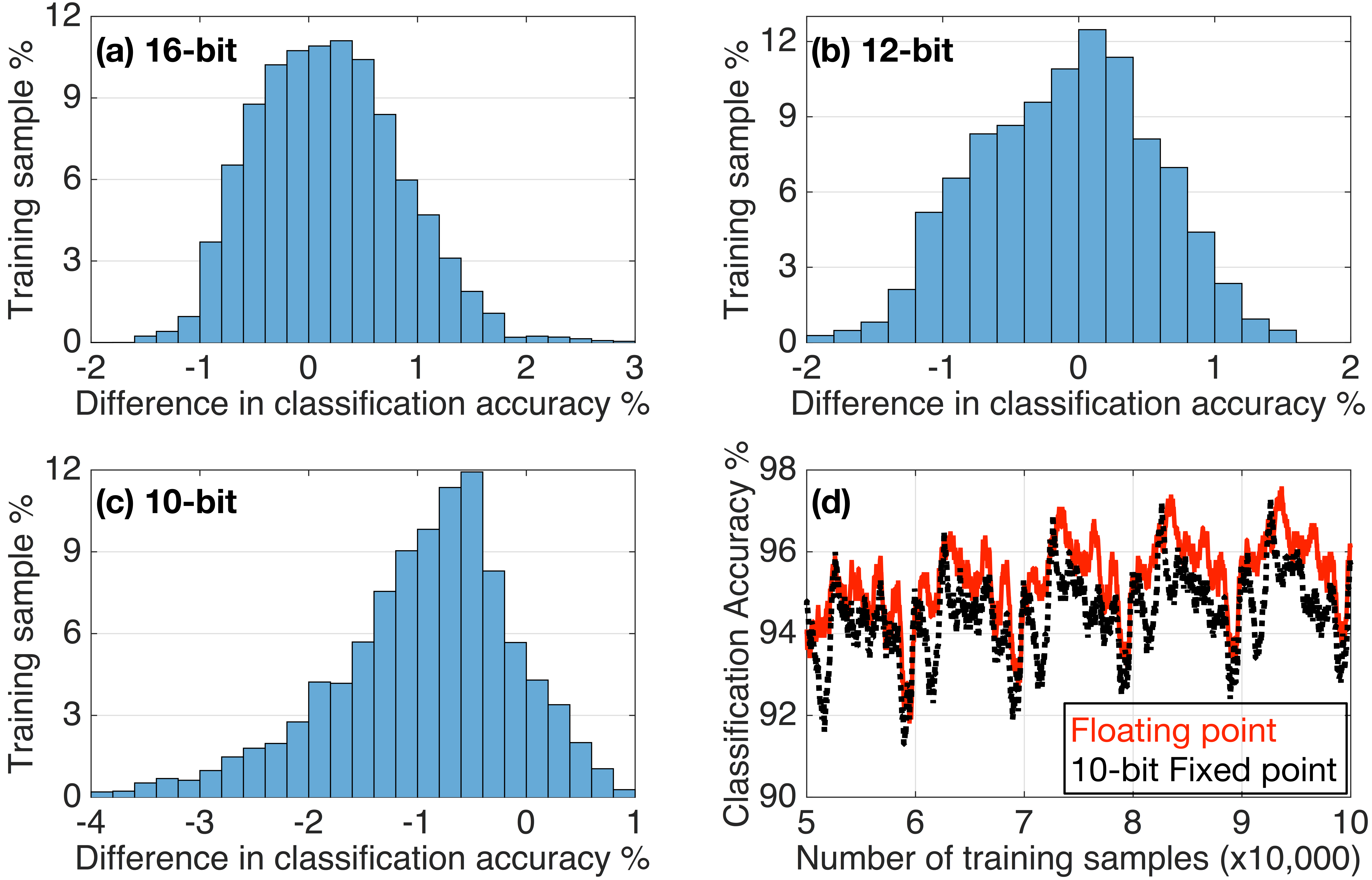}
\caption{\textbf{(a)}, \textbf{(b)}, \textbf{(c)}: Classification accuracy difference histograms for 16-bit, 12-bit and 10-bit fixed point, respectively, with floating point. \textbf{(d)}: Detailed comparison view of a portion of the network learning curves for 10-bit fixed point vs floating point.}
\label{fig-results}
\end{figure}

As proof of concept, we used Verilog Hardware Description Language to develop the register-transfer level (RTL) design for our hardware architecture, and simulated using the MNIST dataset for different fixed point bit widths. 
The neuron configuration is (1024,64,16) (we used powers of 2 for ease of hardware implementation and set the extra neurons to 0) and the fan-out for both junctions is 8, resulting in an 87\% sparse network. The first and second junctions have $z=512$ and $z=32$, respectively. Figures \ref{fig-results}a, \ref{fig-results}b and \ref{fig-results}c show histograms for the classification accuracy difference ``\emph{fixed point} - \emph{floating point}'' (i.e., more bars on the positive side indicate better fixed point performance). Figure \ref{fig-results}d 
indicates that 10-bit fixed point (we used 3 integer bits and 7 fractional bits) for all network parameters and computed values is sufficient to obtain classification performance very close to that of floating point simulations. The plots are for 10,000 training images over 10 epochs.

\section{Conclusion and Future Work}\label{future}
This work presents a flexible architecture that can perform both training and inference of large and deep neural networks on hardware. Sparsity is preset, which greatly reduces the amount of memory and number of multiplier units required. A reconfigurable degree of parallel edge processing enables the architecture to adapt itself to any network size and hardware resource set, while junction pipelining and operational parallelization lead to fast and efficient performance. Our ultimate goal is to propel a paradigm shift from offline training on CPUs and GPUs to online training using the speed and ease of reconfigurability offered by FPGAs and custom chips. Future work would involve extension to other types of networks, tackling memory bandwidth issues, and extensive parameter space exploration to advance the limited theoretical understanding of DNNs.

\bibliographystyle{splncs03}
\bibliography{aaa_main.bib}




























\end{document}